\documentclass[10pt]{article}

\usepackage{mathptmx}

\usepackage{anyfontsize}

\usepackage[a4paper, top=1in, bottom=1.2in, left=1.2in, right=1.2in]{geometry}

\usepackage{setspace}
\usepackage{amsmath, amssymb}
\usepackage{graphicx}
\usepackage{float}
\usepackage{booktabs}
\usepackage{tabularx}
\usepackage{multirow}
\usepackage{threeparttable}
\usepackage{textcomp}
\usepackage{microtype}

\usepackage[english]{babel}
\usepackage{csquotes}
\MakeOuterQuote{"}
\usepackage{upquote}

\usepackage{tipa}

\usepackage[colorlinks=true, citecolor=blue, urlcolor=blue]{hyperref}

\usepackage{orcidlink}

\usepackage{apacite}
\newcommand{\hashtag}[1]{\textit{\##1}}

\title{\textbf{Monsoon Uprising in Bangladesh: How Facebook Shaped Collective Identity}}
\author{
Md Tasin Abir\,\orcidlink{0009-0004-4730-9782} \and
Arpita Chowdhury\,\orcidlink{0009-0003-8531-4788} \and
Ashfia Rahman
}

\date{}

\makeatletter
\renewcommand\@makefnmark{\hbox{\textsuperscript{\normalfont\@thefnmark}}}
\makeatother

\begin{document}

\maketitle

\footnotetext[1]{Emails: \texttt{tasinabir@gmail.com}, \texttt{arpita.chowdhury.edu.bd@gmail.com}, \texttt{rahmanorni7199@gmail.com}. All authors are affiliated with University of Chittagong.}

\begin{abstract}
This study investigates how Facebook shaped collective identity during the July 2024 pro-democracy uprising in Bangladesh, known as the Monsoon Uprising. During government repression, protesters turned to Facebook as a central space for resistance, where multimodal expressions, images, memes, videos, hashtags, and satirical posts played an important role in unifying participants. Using a qualitative approach, this research analyzes visual rhetoric, verbal discourse, and digital irony to reveal how shared symbols, protest art, and slogans built a sense of solidarity. Key elements included the symbolic use of red, the ironic metaphorical use of the term “Razakar,” and the widespread sharing of visuals representing courage, injustice, and resistance. The findings show that the combination of visual and verbal strategies on Facebook not only mobilized public sentiment, but also built a strong collective identity that challenged authoritarian narratives. This study tries to demonstrate how online platforms can serve as powerful tools for identity construction and political mobilization in the digital age.
\end{abstract}

\section{Introduction}
The uprising, which swept across Bangladesh in July-August 2024, also known as the "Monsoon Uprising," earned its name not just from its timing during the peak of the rainy season but also from the powerful symbolism of rain itself. Protesters braved floods and storms, transforming these natural elements into a poignant symbol of resistance and resilience. The term "Monsoon Uprising" thus encapsulates both the literal and metaphorical storms generated by the movement, adding a poetic depth to a defining moment in the nation's history.
This Student-People's Uprising marked a crucial turning point in Bangladesh's political landscape, culminating in the resignation of then Prime Minister Sheikh Hasina. While initially sparked by the reinstatement of a controversial job-quota policy, the movement swiftly broadened, fueled by widespread public discontent over authoritarianism, human rights violations, economic recession, and allegations of rigged elections. In this tumultuous period, social media, particularly Facebook, emerged as a critical facilitator, enabling protesters to organize, exchange vital information, and forge a cohesive collective identity.
This research analyzes how a platform like Facebook helped build a collective identity among protesters by strategically combining visual elements, such as pictures, memes, and videos, with accompanying text. In this digital era, the interplay of visual and verbal rhetoric is crucial for understanding the dynamics and nature of contemporary social movements. This study investigates how Facebook-based discourse, rhetoric, language, and semiotic visualization shaped and expressed collective resistance during this significant social movement.

\section{Literature Review}
The study of social movements has consistently underscored the central role of collective identity in mobilizing individuals and promoting unity. \citeA{melucci1995} defines collective identity as "an interactive and shared definition produced by several individuals (or groups) and concerned with the orientation of action and the field of opportunities and constraints in which the action takes place". It functions as a motivational anchor, binding participants to a shared cause and nurturing the sense of "we-ness" and emotional solidarity essential for sustained activism \cite{polletta2001}.
While classic social movement theories, such as resource mobilization theory \cite{mccarthy1977} and political process theory \cite{tarrow1998}, primarily focus on organizational dynamics and political opportunities, new social movement theory \cite{touraine1985} emphasizes the cultural and identity dimensions of collective action, particularly in post-industrial societies. These theoretical shifts have become even more pronounced with the advent of digital technologies, which facilitate the emergence and flourishing of decentralized, horizontal networks.
Recent studies show how social media platforms, including Facebook, serve as important spaces for identity articulation and meaning-making. \cite{bennett2012} contend that connective action, enabled by digital platforms, allows for personalized engagement in movements even in the absence of formal organizational structures. Consequently, social media transcends its role as a mere logistical tool, becoming a symbolic arena where shared grievances, aspirations, and emotions are visually and verbally encoded to construct collective identity \cite{trere2019}.
The role of visual rhetoric is increasingly essential in the context of the digital landscape. Visual rhetoric refers to how images function persuasively, often employing ethos, pathos, and logos to convey meaning \cite{foss2005}. In the context of social movements, visual elements such as memes, posters, photographs, and videos act as symbolic shorthand, eliciting emotional responses and rallying collective memory. For instance, \citeA{hariman2007} argue that iconic images are instrumental in shaping public consciousness by encapsulating complex political sentiments.
Visuals do not exist in isolation; they interact dynamically with verbal rhetoric, slogans, hashtags, and narratives to create a powerful multimodal rhetoric. Hashtags like \hashtag{RedforJustice} or \hashtag{IAmRazakar} become discursive anchors that unify fragmented voices, transforming individual expression into collective discourse \cite{papacharissi2015}. This symbiosis between the visual and the verbal brings up a potent rhetorical space where identities are not only represented but actively formed.
Irony, especially in digital satire, has emerged as a powerful tool of political subversion. Studies demonstrate that verbal irony, where the literal meaning is inverted, enables protesters to mock power and reclaim derogatory narratives \cite{milner2013}. For example, during the July 2024 Bangladesh uprising, the ironic reclamation of the term "Razakar" transformed a slur into a symbol of unity and defiance, illustrating the protesters' capacity to undermine state narratives through linguistic creativity \cite{zeng2022}.
Artistic expressions, such as political cartoons and protest art, also play a key role in identity formation. Visual artists like Debashish Chakrabarty contributed compelling visual artifacts that circulated widely online, transforming individual creativity into shared symbols of resistance. These artworks resonate deeply within collective memory, serving as digital monuments to the movement’s ideals.
Empirical studies from similar contexts corroborate these insights. For instance, \citeA{ayers2003}, in his study on feminist groups, observed how digital spaces facilitate the negotiation of identity through imagery and dialogue. Similarly, \citeA{trere2016} documented how Mexican student movements skillfully blended humor, art, and rhetoric to form counter-publics.
Within the Bangladeshi context, researchers have explored the crucial role of platforms like Facebook during past movements, such as the Shahbagh protests \cite{islam2015, zaman2016}. The 2024 uprising exemplifies a more evolved phase, where digital rhetoric, semiotic resistance, and emotive storytelling intricately intertwine to forge a potent collective identity in real-time.
Thus, this study builds upon a theoretical foundation, combining social movement theory, digital media studies, visual rhetoric, and critical discourse analysis, to understand how the July 2024 uprising in Bangladesh harnessed the power of Facebook to construct and reinforce collective identity. This literature review offers a strong framework for understanding the strategic use of imagery, satire, and slogans in sustaining political momentum amidst state repression and digital censorship. On social media, the interaction between visual and verbal rhetoric creates a multimodal environment where the combination of text and images produces a synergistic effect. This research analyzes how the July uprising in Bangladesh leveraged this mix of visual and linguistic elements on Facebook to shape collective identity, demonstrating the profound power of rhetoric and semiotic visualization. This study also acknowledges the crucial role of irony in political discourse, where criticism of dominant ideologies is often expressed by conveying meanings that diverge from their literal sense. Furthermore, within social movements, art forms such as cartoons and memes play a significant dual role as both political expression and powerful tools for mass mobilization.

\section{Methodology}
A qualitative approach was used for this study, focusing on the analysis of publicly available Facebook data related to the Bangladeshi uprising in July. The primary data source comprises Facebook posts, encompassing text, images, and videos, along with their associated comments and shares, generated by both participants and observers of the uprising.
Information was collected by identifying relevant content through comprehensive keyword and hashtag analysis. This research targeted terms such as "Bangladesh uprising," "July Revolution Bangladesh," "quota reform movement Bangladesh," and "student protests Bangladesh." It also examined vocabulary related to verbal irony and artistic expressions. Additionally, widely discussed and frequently used hashtags during the movement, including \hashtag{StepDownHasina}, \hashtag{RedForJustice}, \hashtag{JulyMassacre}, and \hashtag{IAmRazakar}, along with other pertinent tags, were considered key data points in the research findings.

\subsection{Data Collection}

The period of data collection spanned from July to the first week of August 2024, aligning with the peak activity of the mass uprising. In this study, both purposive and random sampling methods were judiciously applied. Under purposive sampling, the focus was on high-engagement posts, content from influential protest-related pages and groups, materials containing verbal irony, artwork by artist Debashish Chakrabarty, and widely circulated memes and cartoons. Conversely, the random sampling method was utilized to ensure a broader and more representative inclusion of the diverse range of user-generated content.
Data scraping and archiving were conducted using systematic manual methods, given the constraints of Facebook’s API and the platform’s restrictions on automated tools. Relevant public posts, comments, and discussion threads were identified and collected through direct observation and manual recording, including copy-pasting textual content and capturing screenshots where appropriate. These materials were organized and archived in a secure digital repository for analysis. Ethical considerations have been given the highest priority throughout the research process. Although the data utilized was publicly available, the study has strictly adhered to ethical guidelines, including ensuring confidentiality, anonymization, and responsible use of information. Any personally identifiable information was removed or anonymized during both the analysis and presentation of results.

\subsection{Data Analysis}
The data analysis employed a combination of qualitative methods to comprehensively interpret the collected materials. These methods included qualitative content analysis, which focused on examining textual and discursive elements; visual rhetorical analysis, used to interpret the meaning and rhetorical functions of images and videos; verbal irony analysis; and discourse analysis, which explored how language and visuals interacted to produce meaning.
Based on the research questions and existing literature, coding categories were developed to capture a wide range of visual and emotional elements, with particular attention to ironic expressions and artistic representations. Through this process, recurring themes, metaphors, and rhetorical strategies were identified.
The analysis emphasized the interaction of visual and emotional elements, drawing on tools from discourse and semiotic theory to explore how collective identity was constructed within the digital protest space. In parallel, the study critically examined the impact of government-imposed internet shutdowns as a significant contextual factor, assessing how these disruptions influenced online communication strategies and the evolving dynamics of collective identity formation.

\section{Findings}
This chapter presents the empirical findings of the study, detailing how the collective identity of the Monsoon Uprising was forged through a complex interplay of visual, verbal, and multimodal rhetoric on Facebook. The findings are organized into three main sections: the analysis of standalone visual rhetoric, the examination of unifying verbal discourse, and finally, an exploration of their synergistic combination, which ultimately defined the movement's potent online presence.

\subsection{Visual Rhetoric and Collective Identity Formation}
This section analyzes the visual artifacts that functioned as the primary emotional and symbolic drivers of the movement. In the visually saturated environment of Facebook, images operated as immediate, visceral forms of communication, capable of transcending linguistic barriers and evoking powerful, shared emotional responses. These visuals were not mere illustrations of events; they were active agents in framing the narrative of the uprising and constructing a shared sense of purpose and identity among participants.   

\subsubsection{Iconic Photography and Political Symbolism}
Among the thousands of images that defined the July Uprising, certain photographs achieved iconic status, encapsulating the entire political struggle within a single, emotionally charged frame. The most prominent example was a widely circulated photograph of a student, Abu Sayed, standing alone and unarmed, facing down a formidable line of heavily equipped riot police. 

\begin{figure}[H]
    \centering
    \includegraphics[width=0.50\textwidth,height=0.35\linewidth,keepaspectratio]{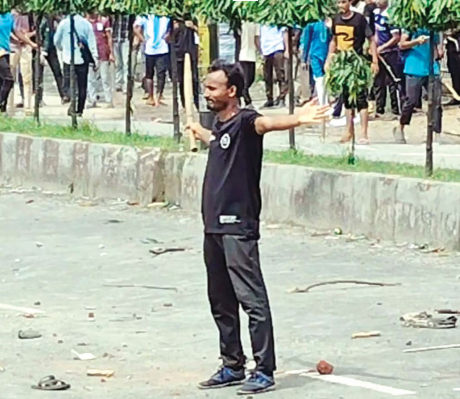}
    \caption{Abu Sayed confronting riot police}
    \label{fig:1_png}
\end{figure}

The rhetorical function of this photograph was to Evoke Courage and Defiance. Its composition, a single, vulnerable individual representing the citizenry against the overwhelming, faceless force of the state, created a powerful David-and-Goliath narrative that resonated deeply with the protesters' sense of moral righteousness. The image transcends simple documentation; it functions as what \citeA{hariman2007} describe as an "iconic image," one that shapes public consciousness by distilling complex political sentiments into a memorable visual form.   
Its contribution to collective identity was profound, serving as a potent Symbol of resistance and shared struggle. The process by which this occurred reveals how collective identity is actively constructed through digital circulation. An individual's act of bravery was captured and shared on Facebook, where the community of protesters, through countless shares, comments, and re-appropriations, collectively designated this image as representative of the entire movement. Consequently, Abu Sayed ceased to be merely one person and was transformed into an avatar for the shared identity of "the protester": brave, defiant, and unyielding. Protesters adopted the image for their profile pictures and banners, performing a visual act of allegiance to the movement’s core values and signaling their own participation in this shared identity.   
\subsubsection{Video Documentation and Emotional Mobilization}
Alongside still images, raw, user-generated video footage played a crucial role in shaping the affective landscape of the uprising. Clips showing police and pro-government activists launching violent attacks on unarmed student protesters circulated widely, serving as crucial evidentiary documents in the face of state-media denial and official propaganda.   

\begin{figure}[H]
    \centering
    \includegraphics[width=0.50\textwidth,height=0.35\linewidth,keepaspectratio]{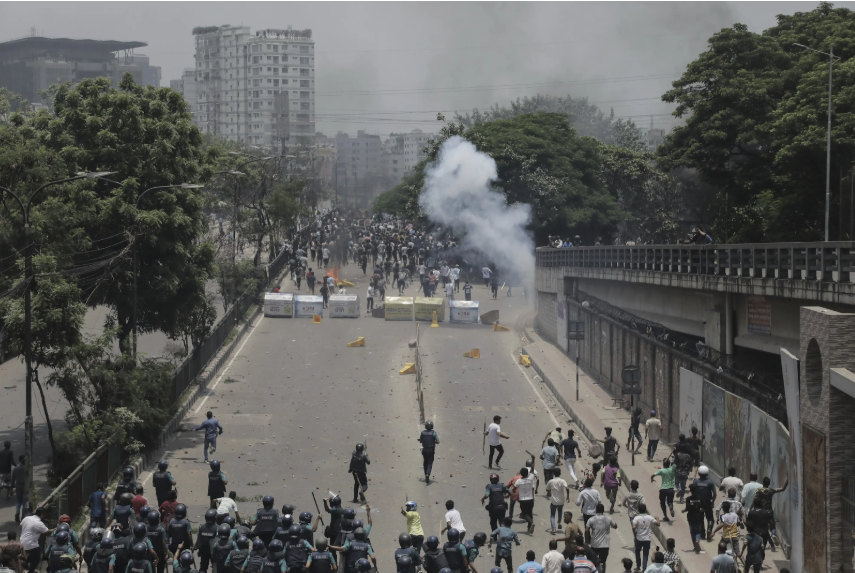}
    \caption{Video stills of police attacks on students}
    \label{fig:2_png}
\end{figure}

The primary rhetorical function of this footage was to Evoke Outrage and Empathy. Unlike static photographs, the visceral nature of video, capturing motion, the sounds of chaos and cries, and the brutal reality of violence, made the state's repression immediate and undeniable. These videos did not allow for passive viewing; they implicated the audience in the unfolding events, generating powerful feelings of anger, sorrow, and moral indignation. This directly contributed to forging a Shared sense of victimhood and solidarity against repression. By witnessing the same brutal events in real-time, a geographically dispersed audience on Facebook became affectively linked. This shared spectatorship of suffering was instrumental in building a cohesive collective identity, transforming abstract grievances into a tangible, emotional experience that created a moral imperative to stand with the victims and oppose the perpetrators.   
\subsubsection{Color Symbolism in Digital Protest}
Digital activism during the uprising was also characterized by simple yet powerful symbolic acts, most notably the widespread adoption of red-colored profile pictures. This practice, often accompanied by hashtags like \hashtag{RedforJustice}, became a ubiquitous visual signature of the movement on Facebook.   

\begin{figure}[H]
    \centering
    \includegraphics[width=0.50\textwidth,height=0.35\linewidth,keepaspectratio]{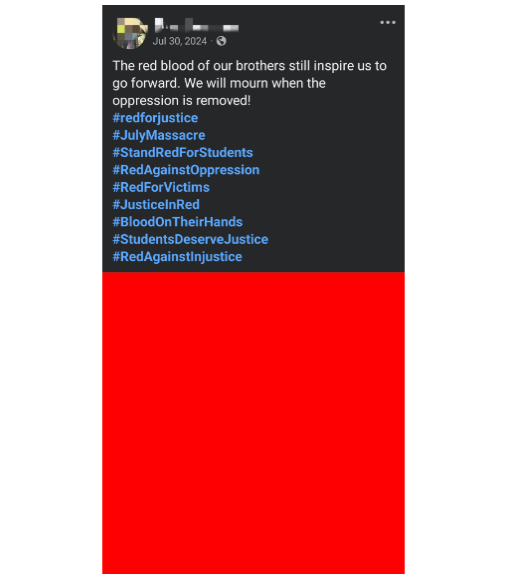}
    \caption{Red-colored profile pictures with hashtags}
    \label{fig:placeholder}
\end{figure}

The rhetorical function of this act was one of Signaling Protest and Solidarity. The color red is a potent and polysemic symbol, simultaneously signifying the blood of martyrs, the sacrifice of protesters, the danger of the political moment, and the revolutionary fervor of the uprising. By overlaying their profile picture with a red filter, users performed a simple semiotic act that made their political allegiance instantly visible to their entire social network. This low-effort, high-visibility form of participation allowed for mass mobilization and expression of unity. The result was a powerful Visual marker of belonging to the movement. Individual Facebook feeds were transformed into a cascading mosaic of dissent, visually demonstrating the vast scale and unity of the opposition and creating a sense of being part of a large, powerful, and committed collective.   
\subsubsection{Political Art as Mobilizing Discourse}
The visual landscape of the protest was further enriched by the contributions of artists who translated the movement's sentiments into compelling visual artifacts. The work of graphic artist Debashish Chakrabarty was particularly prominent, providing the uprising with a distinct aesthetic that elevated protest communication from raw documentation to curated political statement. His posters, which combined stylized imagery with potent slogans, circulated widely as digital manifestos.  

\begin{figure}[H]
    \centering
    \includegraphics[width=0.50\textwidth,height=0.35\linewidth,keepaspectratio]{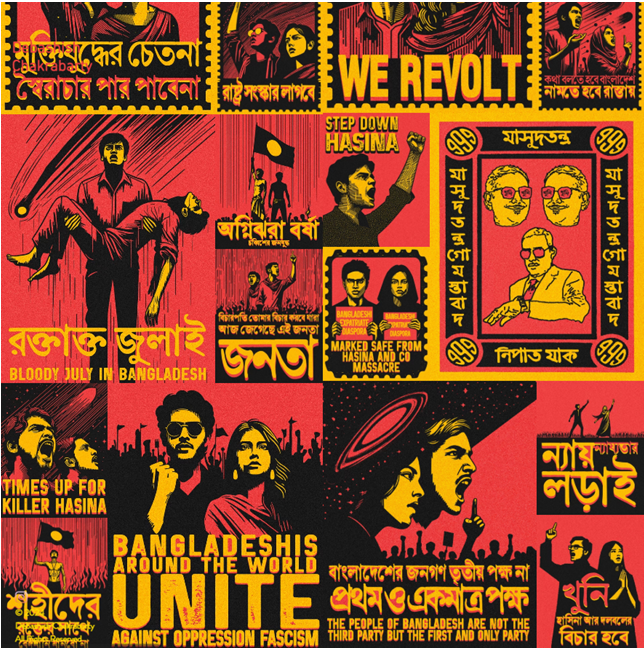}
    \caption{Posters by Debashish Chakrabarty}
    \label{fig:4_png}
\end{figure}

The rhetorical function of this protest art was to Mobilize Resistance. Chakrabarty's posters were not just illustrations; they were calls to action. By fusing compelling graphic design with defiant verbal messages, they created artifacts that were both aesthetically engaging and politically charged. This artistic intervention contributed to a Shared artistic expression of anti-government sentiment. Sharing this art on Facebook became an act of identity curation; it was not merely about spreading a message but also about aligning oneself with a particular cultural and intellectual vision of the resistance, one that was creative, sophisticated, and unflinchingly opposed to authoritarianism. This shared aesthetic sensibility helped forge a more complex collective identity, one rooted not only in grievance but also in cultural defiance.   
\subsubsection{Satirical Visual Media in Political Discourse}
Humor and satire emerged as key weapons in the protesters' rhetorical arsenal, used to puncture the aura of authority and invincibility surrounding the regime. Memes and political cartoons became a primary vehicle for this critique, distilling complex political arguments into easily digestible and highly shareable formats. Examples included a meme showing the Prime Minister painting a pie chart to represent the disproportionate and controversial job quota allocation, and satirical cartoons lampooning the statements and actions of key political figures.

\begin{figure}[H]
    \centering
    \includegraphics[width=0.50\textwidth,height=0.35\linewidth,keepaspectratio]{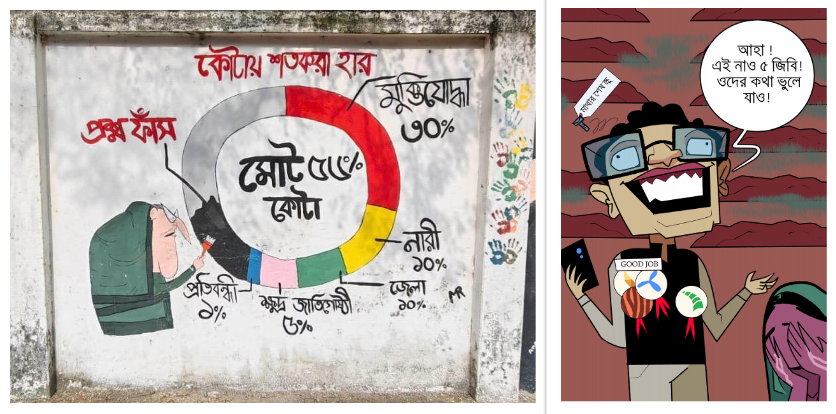}
    \caption{Cartoon lampooning political figures}
    \label{fig:6_png}
\end{figure}

The rhetorical function of these satirical images was twofold: Critiquing Government Policy and providing Political Commentary through Humor. They operated by creating an in-group of those who understood the embedded critique and the cultural references, developing a Shared understanding of grievances and political commentary. This shared humor built solidarity and served as a form of "discursive resistance," allowing protesters to mock, undermine, and de-legitimize power without engaging in direct confrontation, which was often dangerous. The laughter they generated was not frivolous; it was a collective act of defiance that chipped away at the regime's legitimacy and reinforced the protesters' sense of moral and intellectual superiority.   \\
These varied visual forms did not operate in isolation but rather formed a symbiotic rhetorical ecosystem. The process often began with an event of state violence, which was captured in raw, unfiltered videos and photographs, providing the authentic evidence that generated initial outrage and pathos. This raw content was then processed and re-contextualized by the digital community. A moment of individual courage from within the chaos would be isolated and elevated into an iconic symbol of collective defiance. The political grievances underlying the event would then be abstracted and satirized in memes and cartoons, making the critique accessible and shareable. Finally, artists like Chakrabarty would synthesize the raw emotion and political critique into curated, aesthetic statements that served as mobilizing manifestos. This flow from raw documentation to symbolic icon, satirical critique, and artistic expression demonstrates a sophisticated, multi-layered visual strategy that both documented reality and constructed a powerful narrative frame around it, ensuring the movement's message was communicated on multiple emotional and intellectual levels to sustain engagement.

\subsection{Linguistic Tools of Unity and Resistance}

Complementing the powerful visual assault was a range of linguistic and discursive strategies that served as the verbal glue binding individual protesters into a coherent collective. Through slogans, hashtags, and shared narratives, participants on Facebook forged a unified voice, articulating a clear set of grievances, demands, and a shared identity rooted in resistance.   

\subsubsection{Foundational Slogans and Digital Rallying Cries}
At the core of the movement's verbal rhetoric were clear, concise, and powerful slogans that became digital rallying cries. The most prominent of these was the unambiguous demand, "Step Down Hasina," which appeared in countless posts, comments, and image captions. This was consistently paired with the use of inclusive pronouns like "we" and "ours," which linguistically constructed a boundary between the unified protesters and the government.   
This verbal strategy was amplified through the strategic use of hashtags. Tags such as \hashtag{StepdownHasina}, \hashtag{RedForJustice}, and \hashtag{JulyMassacre}, along with more specific ones like \hashtag{RedforVictims}, \hashtag{StudentInRed}, \hashtag{RedAgainstOppression}, and \hashtag{BloodOnTheirHands}, performed a crucial dual function. Their primary rhetorical purpose was Organizing and Signaling Support. These hashtags aggregated thousands of disparate posts into searchable, dynamic archives, creating virtual spaces for discussion and information dissemination. Simultaneously, using a hashtag was an act of public alignment, a way for individuals to signal their support for the movement. Together, these slogans and hashtags created a Unified call for action and shared political objective, facilitating connection and identification among a decentralized network of actors and giving the movement a coherent and powerful voice.   

\subsubsection{Narratives of Shared Struggle and Resilience}
Beyond broad slogans, the collective identity of the movement was significantly deepened through the circulation of personal narratives. Protesters took to Facebook to share their individual stories and testimonials of facing police brutality, arbitrary arrests, and other forms of state-sponsored suppression. These narrative snippets moved the discourse from abstract political claims to the concrete, emotional reality of lived experience.   
The rhetorical function of these posts was the Sharing Experiences of Struggle. By recounting their personal ordeals, protesters developed a profound sense of empathy and connection among the wider audience. These stories of suffering and endurance were instrumental in building collective identity, as they transformed the movement from a political campaign into a community of individuals bound by common hardship and a shared quest for justice. This narrative work was crucial in reinforcing the "us versus them" frame, providing personal, emotional evidence of the state's oppression and solidifying a collective identity based on shared resilience in the face of adversity.   

\subsubsection{Ironic Reappropriation of Political Labels}

Perhaps the single most potent rhetorical event of the uprising was the protesters' ironic reclamation of a deeply pejorative term. In an attempt to delegitimize the movement, Prime Minister Sheikh Hasina referred to the student protesters as "the grandchild of Razakar". The term "Razakar" carries immense historical weight in Bangladesh, referring to the local collaborators who sided with the Pakistani army during the 1971 Liberation War. It is one of the most profound insults in the national lexicon, synonymous with "traitor."   
The state's intent was clear: to frame the protesters as anti-national enemies of the state. However, the movement responded with an act of brilliant "discursive jujitsu." Protesters began to ironically embrace the label, chanting in the streets, "Tumi ke? Ami ke? Razakar, Razakar" (Who are you? Who am I? Razakar, Razakar), and flooding Facebook with the hashtag \hashtag{IAmRazakar}. The rhetorical function of this move was the  Reclaiming of a Derogatory Label. This act of ironic reversal transformed a weapon of shame into a powerful symbol of defiance.

The contribution of this single act to collective identity cannot be overstated. It created a Shared identity through resistance and defiance, uniting protesters against the government's attempt to label them. By collectively and publicly embracing the slur, protesters not only neutralized its power but also forged a new, potent in-group identity. The meaning of "Razakar" was subversively redefined within the context of the movement to mean "one who bravely opposes the current authoritarian regime." This created a clear, emotionally charged boundary between the "we" of the protesters and the "they" of a government that would label its own youth as traitors. The Prime Minister's statement was a profound strategic error; in an attempt to define and divide the movement, the state inadvertently handed it its most effective and unifying identity-building tool, providing a common enemy and a common banner of defiant irony under which disparate factions could unite.   

\subsection{Integrating Visual and Verbal Rhetoric}
The most sophisticated and persuasive rhetorical strategies employed by the movement involved the deliberate fusion of visual and verbal elements. These multimodal artifacts, memes, captioned images, posters, and annotated videos, created layered meanings that were more persuasive, memorable, and resilient than either text or image could be alone. This synergy was central to constructing and reinforcing the collective identity of the uprising.   

\subsubsection{Multimodal Protest Rhetoric}

A common multimodal tactic involved pairing photographic evidence of state violence with sharp, critical captions. For instance, a photograph of a student with visible injuries would be posted on Facebook with an accompanying text such as, "This is the reality of their democracy \textipa{[aha gOnOtOntro]}".  

\begin{figure}[H]
    \centering
    \includegraphics[width=0.50\textwidth,height=0.35\linewidth,keepaspectratio]{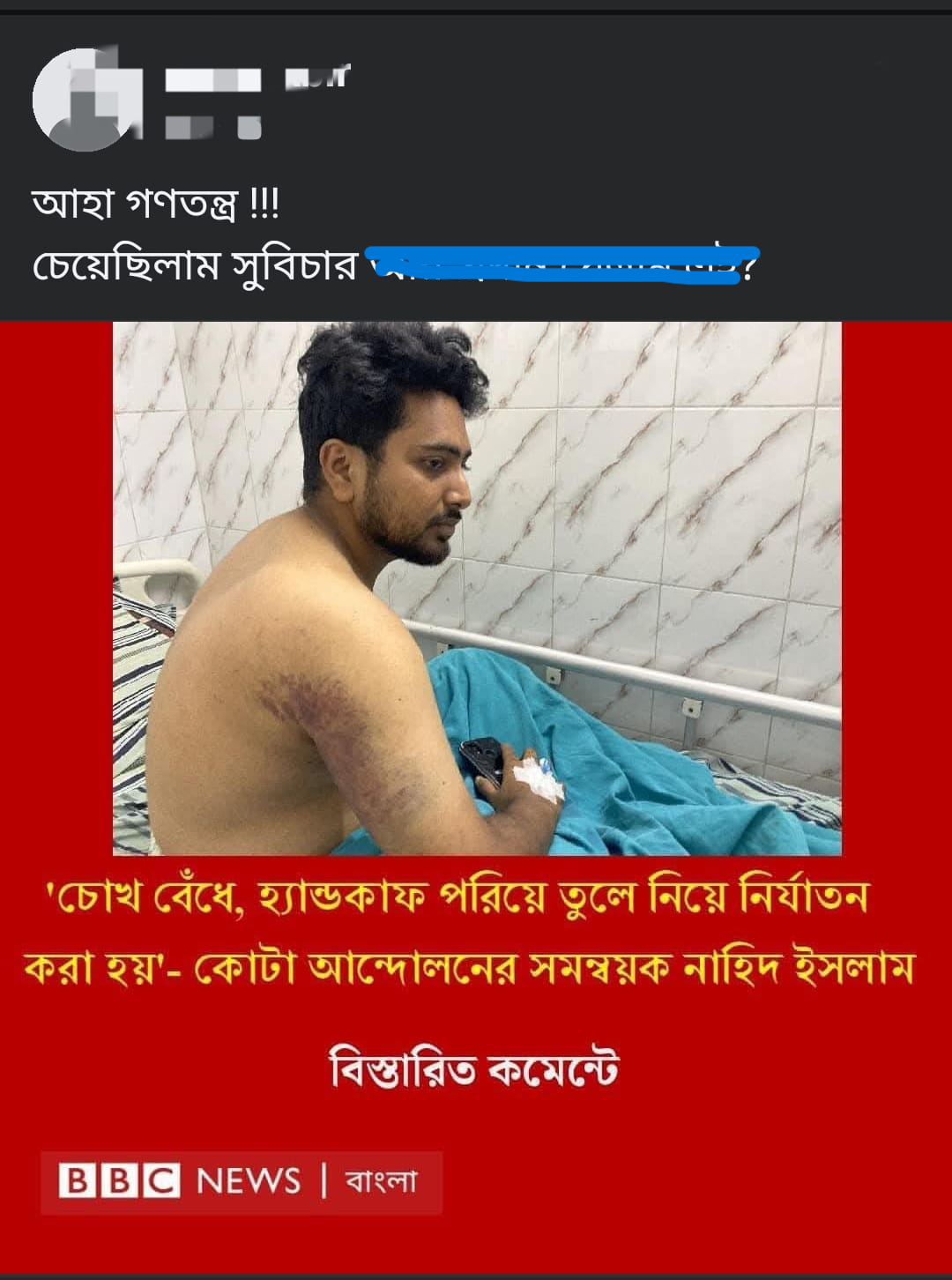}
    \caption{A Facebook post using irony}
    \label{fig:7_png}
\end{figure}

The power of this combination lies in its appeal to both emotion and logic. The image provides the visceral, undeniable evidence of physical harm (pathos), while the text provides the interpretive political frame (logos), explicitly connecting that violence to the government's democratic hypocrisy. This combined rhetoric Reinforces a shared understanding of government actions as oppressive. The juxtaposition forges a sense of shared frustration and righteous anger, strengthening the "us vs. them" dynamic by framing the government not just as violent, but as hypocritical. This created a collective identity based on a shared perception of being victims of a duplicitous regime.   

\subsubsection{Visualizing Injustice Through Metaphorical Memes}
Memes proved to be a particularly efficient form of multimodal rhetoric. An exemplary case is a meme that featured an image of handcuffs with the integrated text, "Our reward for demanding justice".   

\begin{figure}[H]
    \centering
    \includegraphics[width=0.50\textwidth,height=0.35\linewidth,keepaspectratio]{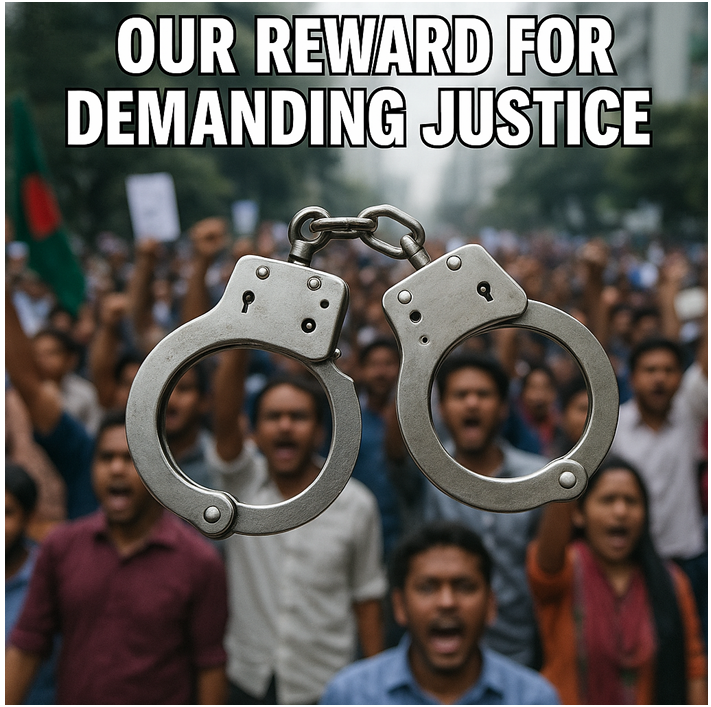}
    \caption{Meme: Metaphorical critique}
    \label{fig:10_png}
\end{figure}

This artifact is a masterclass in rhetorical efficiency. The image of handcuffs serves as a powerful and instantly recognizable visual metaphor for arrest, imprisonment, and repression. The text articulates the protesters' core grievance: the perceived injustice of being punished for exercising their democratic right to protest. The combination Uses visual metaphor to express the perceived injustice of repression, creating a concise, witty, and deeply critical message that is easily shared and understood across the platform. Such memes promoted a sense of shared frustration and cynical solidarity among those who felt punished for seeking their rights, building an identity around a common experience of persecution.   

\subsubsection{Art as Manifesto: The Fusion of Image and Slogan}
The pinnacle of multimodal rhetoric within the movement can be seen in the protest art of figures like Debashish Chakrabarty, where image and text were fused into a singular, powerful statement. A poster combining a stylized, evocative visual with a poetic and aggressive slogan like, "On spring nights like this, I spit in the face of fascists," functions as a digital manifesto.   
\begin{figure}[H]
    \centering
    \includegraphics[width=0.50\textwidth,height=0.35\linewidth,keepaspectratio]{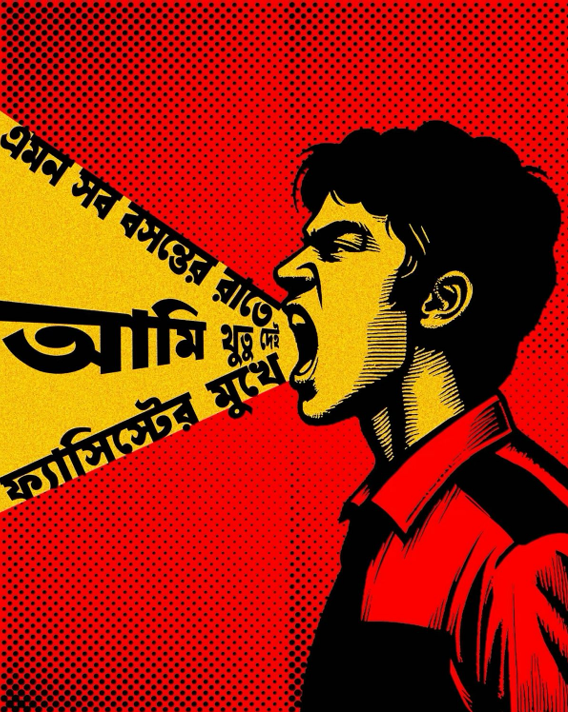}
    \caption{Artistic poster declaring "I spit in the face of fascists"}
    \label{fig:8_png}
\end{figure}
Here, the combination of a high-art aesthetic with a raw, defiant verbal message Amplifies anti-government sentiment and encourages resistance. It moves beyond simple commentary to become a declaration of intent and a cultural touchstone for the movement. The fusion of sophisticated artistry with visceral political anger creates a deeply resonant artifact that both inspires and emboldens the collective. Sharing such a piece was an act of aligning with an identity that was not only politically defiant but also culturally and artistically potent. 

\subsubsection{Satire as Subversive Exposé}
Satirical cartoons offered another potent form of multimodal critique. A typical example would feature a caricature of a politician making a demonstrably absurd statement, with the statement itself quoted in a caption or speech bubble.   

\begin{figure}[H]
    \centering
    \includegraphics[width=0.50\textwidth,height=0.35\linewidth,keepaspectratio]{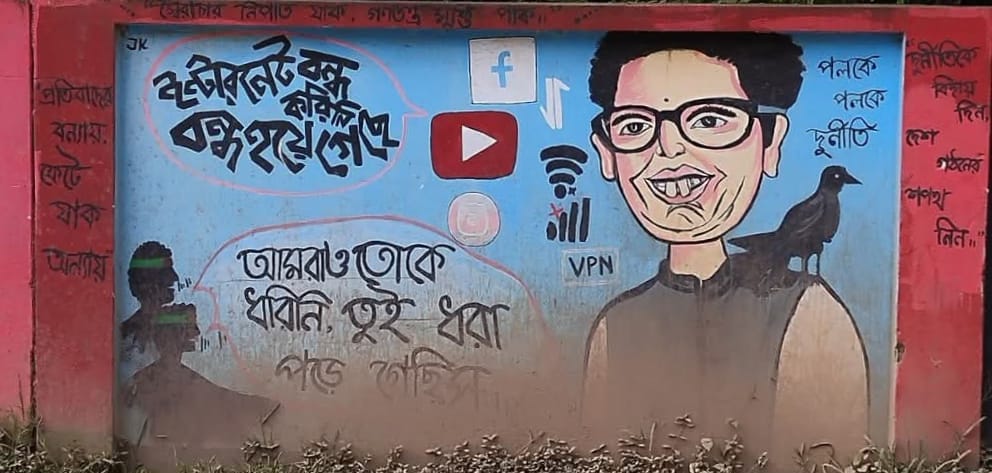}
    \caption{Satirical cartoon of a political figure}
    \label{fig:9_png}
\end{figure}

This format Uses satire to highlight the perceived foolishness of the ruling power. The cartoon visualizes the politician as a ridiculous figure, stripping them of their authority, while the quoted text provides the direct evidence of their perceived incompetence or absurdity. This combination creates shared amusement among the audience and reinforces a critical view of the government, solidifying a collective identity based on a shared sense of intellectual and moral superiority over a regime portrayed as foolish and out of touch.   
\\
The prevalence of these multimodal artifacts was not merely an aesthetic choice but a highly strategic one, particularly in a context of censorship and information warfare. As the government employed tactics like internet shutdowns, efficient communication became paramount. A single multimodal artifact, such as a meme or a poster, is an information-dense package that conveys emotion, argument, and context simultaneously, far more efficiently than a separate image and a long text post. These artifacts are also more resilient to automated censorship, which can easily flag keywords but struggles to interpret the subtle, combined meaning of an image and its embedded text. Furthermore, this multimodality can penetrate cognitive defenses. A regime supporter might dismiss a slogan, but an image of violence paired with a critical caption creates cognitive dissonance, making the protest narrative harder to ignore. Therefore, the movement's multimodal strategy was crucial for both building in-group solidarity and effectively challenging the narratives of the out-group.

\section{Discussion and Conclusion}

\section{Discussion}

The findings of this study illuminate the complex and dynamic processes through which collective identity was constructed and reinforced during the July 2024 Monsoon Uprising in Bangladesh. The analysis demonstrates that Facebook functioned not merely as a logistical tool for coordination but as a vital symbolic arena where a potent sense of shared purpose, grievance, and solidarity was meticulously forged. This discussion interprets the empirical findings by placing them in direct conversation with the theoretical frameworks of social movement studies, visual rhetoric, and digital activism. It argues that the movement’s success in building a cohesive collective identity stemmed from a sophisticated and synergistic deployment of multimodal rhetoric, the cultivation of affective publics, and a masterful use of discursive irony to counter state narratives.

This chapter will proceed in three parts. First, it will analyze the semiotic architecture of the uprising, exploring how a multimodal ecosystem of iconic imagery, raw video footage, political art, and satire created a coherent and emotionally resonant narrative of defiance. Second, it will examine the movement's evolution from the loose, personalized networks characteristic of "connective action" toward a more robust form of "affective-collective action," driven by a "logic of collection" that aggregated individual acts of allegiance into a powerful display of unity. Finally, it will dissect the single most potent rhetorical event of the uprising, the ironic reclamation of the term "Razakar", as a masterful act of discursive reversal that solidified the movement's identity.

\subsection{Semiotic Architecture of Digital Resistance}

The collective identity of the Monsoon Uprising was not a spontaneous phenomenon but was actively and strategically constructed through a sophisticated semiotic architecture on Facebook. This architecture relied on the powerful fusion of visual and verbal rhetoric to create a coherent, emotionally resonant, and politically persuasive narrative of defiance against an authoritarian regime. This process moved beyond simple information dissemination to become an act of collective meaning-making, where digital artifacts were not just shared but were imbued with symbolic power by the community that circulated them.

\subsubsection{Iconic Imagery and Symbolic Defiance}

The foundation of this semiotic architecture was built upon iconic imagery. In their definitive study, \citeA{hariman2007} argue that certain photographs transcend their documentary function to become "iconic images", photographs that are widely recognized and remembered, represent historically significant events, and elicit strong emotional responses. These images function as a dynamic form of public art, serving as signposts for collective memory and resources for critical reflection. They distill complex political sentiments into a single, memorable frame, transforming a specific event into a structure for political identity. The widely circulated photograph of the student Abu Sayed standing alone and unarmed against a formidable line of heavily armed riot police serves as a prime example of this phenomenon.

The rhetorical power of this photograph lies in its composition, which creates a powerful David-and-Goliath narrative, framing the protesters' cause in stark moral terms. The lone, vulnerable individual represents the citizenry, while the heavily equipped, faceless police line represents the overwhelming and oppressive force of the state. This visual trope is a recurring feature in global protest culture, echoing images like the "Tank Man" in Tiananmen Square or the "Woman in Red" from Turkey's 2013 Gezi Park protests, who similarly became international symbols of individual courage against state power \cite{tarrow1998}. The image of Abu Sayed, therefore, did not require a caption; its meaning was immediately legible within this established visual lexicon of dissent \cite{hariman2007}.

However, the photo's iconic status was not inherent in the image itself but was conferred upon it by the digital community. Its power was actualized through the performative act of thousands of users adopting it as their profile picture. This act was not passive sharing; it was a public declaration of allegiance and identification. Through this digital ritual, an individual’s act of courage was collectively designated as representative of the entire movement. Consequently, Abu Sayed ceased to be merely one person and was transformed into an avatar for the shared identity of "the protester": brave, defiant, and unyielding. This process reveals a crucial dynamic of digital activism: collective identity is not a static quality to be discovered but is actively and continuously constructed through the circulation and ritualistic appropriation of shared symbols. The image becomes iconic because the community, through its actions, makes it so, weaving it into the very fabric of their collective self-conception \cite{melucci1995, polletta2001}.

\subsubsection{User-Generated Footage and Affective Mobilization}

This symbolic crystallization of defiance was fueled by the circulation of raw, user-generated video footage depicting state violence. Clips of police and pro-government activists launching brutal attacks on unarmed student protesters served as visceral, undeniable evidence that bypassed the censorship and propaganda of state-controlled media. This finding aligns with the role of social media in the Arab Spring, where platforms like Facebook and YouTube were used to disseminate information and awaken populations to the realities of repressive regimes, creating an alternative evidentiary record that state media could not control \cite{gerbaudo2012}.

The significance of this footage is best understood through the theory of 'affective publics' by \citeA{papacharissi2015}, he argues that in the digital age, publics are increasingly formed and mobilized not through rational deliberation alone, but through shared expressions of sentiment and emotion. Affective publics are "networked public formations that are mobilized and connected or disconnected through expressions of sentiment". They emerge when a story, image, or event generates a powerful emotional current that flows through a network, binding individuals together through a shared feeling.

The visceral nature of the videos from the Monsoon Uprising, capturing motion, the sounds of chaos, and the brutal reality of repression, generated powerful feelings of anger, sorrow, and moral indignation among viewers. This shared spectatorship of suffering was instrumental in creating an affective public. A geographically dispersed audience on Facebook, by witnessing the same brutal events, became emotionally and morally linked. This process transforms abstract political grievances into a tangible, shared experience of victimhood and solidarity, creating a moral imperative to oppose the regime\cite{papacharissi2015}. The affective public thus becomes a potent counter-public, one where the emotional truth of witnessed violence overrides the "official truth" of state propaganda. The bonds forged through this shared affect are a powerful component of collective identity, creating a sense of "we-ness" rooted in a common moral and emotional landscape.

\subsubsection{The Multimodal Rhetoric of Resistance}

The movement’s visual landscape was further enriched by political art and satirical media, which processed the raw emotion of events into more curated and critical forms. The work of graphic artist Debashish Chakrabarty, whose posters combined stylized imagery with potent slogans, functioned as digital manifestos, elevating the movement's communication from raw documentation to an aestheticized call to action. Sharing this art became an act of identity curation, aligning oneself with a vision of resistance that was not only defiant but also creative and culturally sophisticated. This corroborates findings on the crucial role of artists' collectives in mobilizing and sustaining uprisings in contexts like Sri Lanka and Hong Kong, where art helps to shape and preserve the movement's culture.

Simultaneously, satirical memes and cartoons functioned as a key form of "discursive resistance". Memes critiquing the controversial job quota policy or lampooning political figures used humor to de-legitimize authority and puncture the regime's aura of invincibility. This use of subversive humor is a well-documented tactic in authoritarian contexts, allowing protesters to mock power, promote an in-group understanding, and build solidarity without engaging in direct, and often dangerous, confrontation \cite{zeng2022}. The laughter generated by these memes was not frivolous; it was a collective act of defiance that reinforced the protesters' sense of moral and intellectual superiority over a regime portrayed as foolish and out of touch. Similar uses of satire and humor were prevalent in the Arab Spring and the Gezi Park protests, where wit and mockery became vital tools for challenging power and building morale \cite{gerbaudo2012}.

These varied visual forms did not operate in isolation but constituted a sophisticated, multi-layered rhetorical ecosystem, a kind of "rhetorical supply chain" for meaning-making. The process often began with an event of state violence, which was captured in raw video and photographs, providing the authentic evidence that generated initial outrage and pathos. From this chaos, a moment of individual courage was isolated and elevated into an iconic symbol of collective defiance, providing ethos. The underlying political grievances were then abstracted and critiqued through accessible and shareable satirical memes, providing logos and satire. Finally, artists synthesized the raw emotion and political critique into curated, aesthetic manifestos that served as mobilizing calls to action. This progression from raw documentation to symbolic icon, satirical critique, and artistic expression reveals a highly adaptive communication strategy that engaged audiences on multiple emotional and intellectual levels, ensuring the movement's narrative was both powerful and resilient.

\subsection{From Connective to Affective-Collective Action}

While the Monsoon Uprising exhibited features of what \citeA{bennett2012} term "connective action", mobilization based on personalized expression and the sharing of content across digital networks, its true strength lay in its ability to transcend the loose, ephemeral connections that often characterize such movements and forge a robust, durable collective identity. In the logic of connective action, the emphasis is on the individual's personalized framing of an issue, and collective identity is seen as less central than in traditional social movements. However, the dynamics observed during the Monsoon Uprising suggest a more complex process was at play, one that points toward the construction of a strong collective identity, not its diminishment.

To understand this transition, it is useful to introduce the concept of the 'logic of collection' by \citeA{gerbaudo2012} as a crucial extension or counterpoint to connective action. He argues that the primary goal of many digital campaigns, particularly those involving personal testimony or symbolic acts, is not merely to "connect" individuals in a network but to "collect" their expressions under a shared banner. This process involves gathering and aggregating similar individual acts into a visible, quantitative whole. The power of these campaigns derives from this accumulation, which serves a demonstrative purpose by highlighting the sheer scale of a grievance and constructing a sense of "we-ness" from the bottom up \cite{gerbaudo2012}.

The widespread adoption of red-colored profile pictures, often accompanied by hashtags like \hashtag{RedForJustice}, is a prime example of this "logic of collection" in action. This simple, low-effort, yet highly visible semiotic act transformed individual Facebook feeds into a cascading mosaic of dissent. It was a powerful visual marker of belonging, instantly signaling political allegiance to one's entire social network and visually demonstrating the vast scale and unity of the opposition. Similarly, the circulation of personal narratives of struggle, testimonials of facing police brutality or arbitrary arrest, moved the discourse from abstract political claims to the concrete, emotional reality of lived experience. These stories created profound empathy and transformed the movement into a community bound by common hardship. The act of using the hashtag or changing a profile picture becomes an act of "counting oneself in," adding one's voice to an expanding chorus of dissent. This process demonstrates a crucial evolution from "connective action" to what can be termed "affective-collective action." The primary logic was not just the sharing of personalized content but the aggregation of symbolic allegiance. This collective performance, fueled by shared emotions of outrage and solidarity, creates a much stronger bond than simple network connections. It forges a shared identity rooted in a public and performative declaration of belonging. This is particularly potent for emergent groups like the student protesters, as research shows that the link between social identification and participation in collective action is significantly stronger for emergent groups than for pre-existing ones. The movement's success, therefore, was not just in its ability to connect people, but in its ability to collect them into a visible, affective, and unified political body online, thereby building the "we-ness" that \citeA{polletta2001} identify as essential for sustained activism.

\subsection{Discursive Reversal and Identity Solidification}

Perhaps the single most potent rhetorical event of the uprising was the protesters' ironic reclamation of the term "Razakar." This act functioned as a masterful instance of "discursive jujitsu," a rhetorical maneuver that neutralized a state-sponsored attempt at stigmatization and, in doing so, forged the movement's most powerful and enduring identity marker.

The event began when Prime Minister Sheikh Hasina, in an attempt to delegitimize the movement, referred to the student protesters as the "grandchildren of Razakar". In the Bangladeshi context, "Razakar," referring to the local collaborators with the Pakistani army during the 1971 Liberation War, is one of the most profound insults in the national lexicon, synonymous with "traitor". The state's intent was clear: to frame the protesters as anti-national enemies, thereby dividing them and eroding public support.

The movement's response was a stroke of rhetorical brilliance. Instead of rejecting the label, protesters ironically embraced it, flooding Facebook with the hashtag \hashtag{IAmRazakar} and chanting in the streets, "Tumi ke? Ami ke? Razakar, Razakar" (Who are you? Who am I? Razakar, Razakar). This act of ironic reversal transformed a weapon of shame into a badge of defiance. It is a classic example of reframing, where a group accepts a derogatory label but inverts its moral valence, re-signifying "Razakar" within the context of the movement to mean "one who bravely opposes the current authoritarian regime." This strategy aligns with scholarship on the power of political irony and satire to subvert state narratives, mock authority, and forge clear in-group/out-group boundaries \cite{zeng2022}.

This tactic of reappropriating a pejorative term is a recurring pattern in 21st-century protests. During Turkey's 2013 Gezi Park protests, for instance, then-Prime Minister Erdoğan dismissed the demonstrators as "çapulcu" (looters). Protesters immediately embraced the label, with "chapulling" becoming a neologism for "fighting for one's rights" that spread globally \cite{tarrow1998}. In both the Bangladeshi and Turkish cases, the state's attempt to define and marginalize the movement backfired spectacularly, providing activists with a powerful, unifying symbol of their opposition.

The \hashtag{IAmRazakar} moment was not merely a clever tactic; it was a foundational identity-building act that provided the movement with three critical resources simultaneously. First, it solidified a common enemy. The Prime Minister's statement cast the government as an entity so disconnected from its people that it would brand its own youth as traitors, creating a clear "us versus them" dynamic. Second, it forged a shared identity. The hashtag  became a simple, powerful, and defiant banner under which disparate factions could unite. To use it was to declare oneself part of the resistant "we." Third, it secured a crucial moral victory. The state's most potent rhetorical weapon was not only neutralized but turned against it, exposing the regime's desperation and energizing participants with a sense of intellectual and moral superiority. In a profound strategic error, the state, in its attempt to define and divide the movement, inadvertently handed it its most effective and unifying identity-building tool.

\subsection{Conclusion}

This study examined how Facebook played a crucial role in shaping and reinforcing collective identity during the July 2024 Monsoon Uprising in Bangladesh. The findings suggest that the movement’s strength did not lie solely in its demands or the number of participants, but in the strategic use of multimodal rhetoric within the digital public sphere. Through iconic images, gripping video footage, political art, and the ironic reclamation of the term “Razakar,” a decentralized network of individuals transformed into a powerful political force capable of toppling an entrenched authoritarian regime.

The fall of a long-standing government, however, is not an endpoint but a turning point. The real challenge begins with translating the unity and energy of the uprising into the difficult tasks of governance and institutional reform. One notable step has been the formation of the Jatiya Nagorik Party (NCP), created by the student leaders who drove the movement. Emerging from initiatives like the Jatiya Nagorik Committee, the NCP seeks to build a “Second Republic” by rewriting the constitution and building a political culture grounded in merit and justice, rather than dynastic control and retribution.

This transition from protest to political party captures the central question for Bangladesh’s future: can a movement born of digital activism and symbolic resistance create the lasting institutions needed for a stable democracy? The path ahead is full of both promise and risk. Yet, the Monsoon Uprising has shown the extraordinary power of a unified citizenry. Having cast off an authoritarian government, the people of Bangladesh now hold a collective hope for a democratic, inclusive, and just society, one that was first imagined online and voiced in the streets.

While this study centers on Facebook because of its pivotal role in the Monsoon Uprising, it does not capture the impact of other encrypted platforms such as WhatsApp or Telegram, nor the offline organizing that complemented digital activism. Establishing a direct causal link between online rhetoric and mass mobilization also remains a persistent challenge in social science. Despite these limitations, the findings expose a critical gap in existing scholarship and open new directions for research, revealing both the power and fragility of networked movements, their ability to forge collective identities and disrupt political systems, alongside their vulnerability in the uncertain aftermath of upheaval.

The Monsoon Uprising stands as a defining example of how political struggles are evolving in the digital age. It shows that building collective identity is no longer just a precondition for action; it is itself a form of resistance, forged in real time through images, words, and symbols circulating across networked platforms. The mastery of multimodal rhetoric, the mobilization of affective publics, and the subversive use of irony have become essential tools for challenging entrenched power. While the long-term outcome of Bangladesh’s transformation remains uncertain, one lesson is clear: the fight for democracy is increasingly a battle over meaning, waged not only on the streets but also in the symbolic spaces of the digital sphere, where the power to define reality is the ultimate prize. Understanding these dynamics is vital for navigating the shifting political landscapes of the 21st century.
\bibliography{Monsoon_Uprising_in_Bangladesh_How_Facebook_Shaped_Collective_Identity}
\end{document}